\pdfoutput=1
\documentclass[10pt,twocolumn]{ICCAS}

\usepackage[T1]{fontenc}

\usepackage{url}
\usepackage{amssymb}

\graphicspath{{./}}

\makeatletter

\renewcommand{\@seccntformat}[1]{%
  \csname the#1\endcsname
  \ifnum\pdfstrcmp{#1}{section}=0 .\fi
  \quad
}

\def\abbreviations#1{\gdef\@abbreviations{#1}}
\gdef\@abbreviations{}
\def\@maketitle{\newpage \vskip-4mm
\vskip2.0em
      {\lineskip0.5em\center\fontsize{12pt}{12pt}\selectfont\bf\textcolor[RGB]{0,112,192}\@title\par}%
      {\lineskip.5em\center\fontsize{11pt}{11pt}\selectfont\@author\par}
      {\lineskip.5em\center\normalsize\@affils\par}
      \noindent{\vskip 1em \normalsize{\bf \textcolor[RGB]{227,108,10}{Abbreviations:} }\@abbreviations\par}
      \noindent{\vskip 1em \normalsize{\bf \textcolor[RGB]{227,108,10}{Abstract:} }\@abstract\par}
      \noindent{\vskip 1em \normalsize {\bf \textcolor[RGB]{227,108,10}{Keywords:} }\@keywords\par}
    \par\vskip 2pc}
\makeatother

\newcommand{\eps}{\varepsilon}
\DeclareMathOperator{\sat}{sat}

\begin{document}

\setlength{\abovedisplayskip}{4pt plus 2pt minus 2pt}
\setlength{\belowdisplayskip}{4pt plus 2pt minus 2pt}
\setlength{\abovedisplayshortskip}{2pt plus 1pt}
\setlength{\belowdisplayshortskip}{2pt plus 1pt}

\title{The Effectiveness of Sliding Mode Control for Robust\\
Vision Stabilization under the Plant-Binding Condition}

\author{Sooyeun Yang${}^{1,2}$, Jongseong Brad Choi${}^{1,2,3}$ and Seung Bok Choi${}^{1,2*}$}

\affils{${}^{1}$Department of Mechanical Engineering,
SUNY Korea, Incheon, Republic of Korea \\
${}^2$Department of Mechanical Engineering, Stony Brook University, NY, USA \\
${}^3$Department of Computer Science, SUNY Korea, Incheon, Republic of Korea \\
{\small ${}^{*}$ Corresponding author}
}

\abbreviations{
CI: Confidence Interval;
FOV: Field of View;
LOO-CV: Leave-One-Out Cross-Validation;
LPF: Low-Pass Filter;
OIS: Optical Image Stabilization;
ORB: Oriented FAST and Rotated BRIEF;
PID: Proportional--Integral--Derivative;
PTZ: Pan--Tilt--Zoom;
RMS: Root Mean Square;
SMC: Sliding Mode Control.
}

\abstract{
Boundary-layer Sliding Mode Control (SMC) has been applied to vision-based
stabilization, but identifying its \emph{activation condition} against
simpler causal baselines (LPF, PID) has been elusive. The operational
condition is \emph{plant binding}: the disturbance velocity approaches
or exceeds the actuator slew limit, so the bounded per-step actuation
budget binds. We validate this
in two independent vision domains: handheld translation stabilization
($12$ $1080$p clips; LPF, PID, SMC, $L^1$) and angular re-framing
against a virtual slew-rate-limited pan--tilt--zoom (PTZ) plant. Outside binding,
leave-one-out cross-validation makes SMC and LPF on translation
statistically indistinguishable, and PID beats SMC on the angular
$99$\textsuperscript{th}-percentile pointing offset by $5.85^\circ$
(95\% CI clear of zero); inside binding --- reached only by sweeping the
simulated slew limit, as both captured datasets land outside it --- the
translation gap stays a tie
and SMC beats PID on the angular worst case (Cohen's $d$ up to $+0.83$).
The PTZ plant's $100^\circ/$s default sits on the cross-over,
which explains a small-margin PID result previously reported
in the angular setting. A complementary ablation shows raising the SMC reaching gain
$k$ costs mean tracking outside a narrow window.
}


\maketitle

\section{Introduction}\label{sec:intro}

Boundary-layer Sliding Mode Control
(SMC)~\cite{Utkin1977,Slotine1991} is a natural candidate for
vision-based stabilization: the per-frame stabilizing warp can be cast
as a virtual unit-mass plant disturbed by an unknown but bounded shake
signal, and the boundary-layer reaching law of Slotine and
Li~\cite{Slotine1991,SlotineSastry1983} drives the sliding variable
into a tunable layer
in finite time. Yet across recent practical work, SMC variants have
produced mixed and often negative comparisons against simpler causal
baselines such as causal low-pass filtering (LPF) and PID. Our own
prior single-domain experiments are no exception: handheld translation
stabilization placed SMC behind LPF by $2.67$~px above-cutoff residual
on $11$ test clips, and active angular re-framing placed SMC behind PID
by $5.85^\circ$ at the $99$\textsuperscript{th}-percentile pointing
offset on a virtual PTZ plant. Both first-draft conclusions read as
``SMC under-performs'' without identifying the regime in which the
bounded-disturbance robustness actually becomes load-bearing.

An explanation draws on three separate literatures. SMC's
robustness to bounded matched disturbances is
classical~\cite{Utkin1977,Edwards1998}, and practical deployment
guides caution that the discontinuous machinery pays for
itself only against a genuinely binding
uncertainty~\cite{Young1999}. Software video stabilization spans
causal filtering through offline optimal-path
methods~\cite{Grundmann2011,Liu2013,Guilluy2021}, but comparisons
run on an unconstrained (software-warp) actuator. PTZ platforms
carry an explicit mechanical pan--tilt rate
limit~\cite{Sinha2006}, and the saturation literature treats
bounded actuation as the central design
obstacle~\cite{Bernstein1995,Galeani2009}. To our knowledge, the
point at which that constraint \emph{activates} SMC's advantage over
simpler laws in a vision loop has not been isolated experimentally.

This paper resolves both negatives with a single observation. The
boundary-layer machinery activates only when the disturbance velocity
approaches or exceeds the actuator's slew limit --- i.e.\ when a
bounded per-step actuation budget binds against a near-binding
disturbance. We call this the \emph{plant-binding} regime and show
across two independent vision-control domains that it is the
operational \emph{activation condition} for boundary-layer SMC.
Fig.~\ref{fig:block} (Sec.~\ref{sec:model}) summarizes the unified
per-axis control loop in which the claim is tested.

\noindent\textbf{Primary contributions.}
(1)~\emph{An operational binding condition for boundary-layer SMC on
vision-control tasks.}
We identify and empirically validate the binding regime: it is
stated as a normalised binding ratio (Sec.~\ref{sec:model}) and
verified directly by sweeping the plant slew limit through the
crossing in both domains (Sec.~\ref{sec:results:theorem}); the sweep
is simulated, and Sec.~\ref{sec:limitations} scopes the claim
accordingly.
(2)~\emph{Robust vision stabilization inside binding.}
Once the plant binds, boundary-layer SMC uses its
provable insensitivity to the matched shake disturbance and to plant
uncertainty to improve worst-case stabilization over
PID (Cohen's $d$ up to $+0.83$) while staying at least as good on the
mean. On the present simulation evidence, SMC is therefore the
stronger controller inside this actuation-constrained regime; the
hardware demonstration is future work.
(3)~\emph{A two-domain unification.}
The same binding condition governs SMC activation in translation
stabilization and in angular re-framing against a slew-limited plant.
The $100^\circ/$s default of the angular PTZ plant sits
on the cross-over, so earlier draft comparisons that read SMC as
``no better than PID'' were taken at an unbinding
operating point, not a structural SMC deficiency.

\noindent\textbf{Supporting contributions.}
(4)~\emph{A disclosed bandwidth Pareto for the translation
$\sigma_{ty}$ metric.}
SMC and LPF tie under leave-one-out cross-validation ($+0.050$~px,
CI crosses zero); the residual gap at the conservative operating
point $c{=}25$ is a bandwidth choice
(Sec.~\ref{sec:results:pareto}), not an SMC deficiency.
(5)~\emph{A ``higher gain is not free'' ablation.}
We confirm the binding gap is closed by the plant, not by gain
inflation: pushing the SMC reaching gain $k$ from $50$ to $400$
over-actuates and gives no broad improvement, so the tuned
boundary-layer SMC is already the right robust design.

Sec.~\ref{sec:model} unifies the two-domain model and the
boundary-layer design; Sec.~\ref{sec:setup} presents the harness,
the disclosed Pareto trade-off, and the binding cross-over;
Sec.~\ref{sec:ablation} gives the ablations and limitations;
Sec.~\ref{sec:conclusion} concludes.

\section{System Model and Controller Design}\label{sec:model}

\begin{figure*}[t]
\centering
\includegraphics[width=0.66\textwidth]{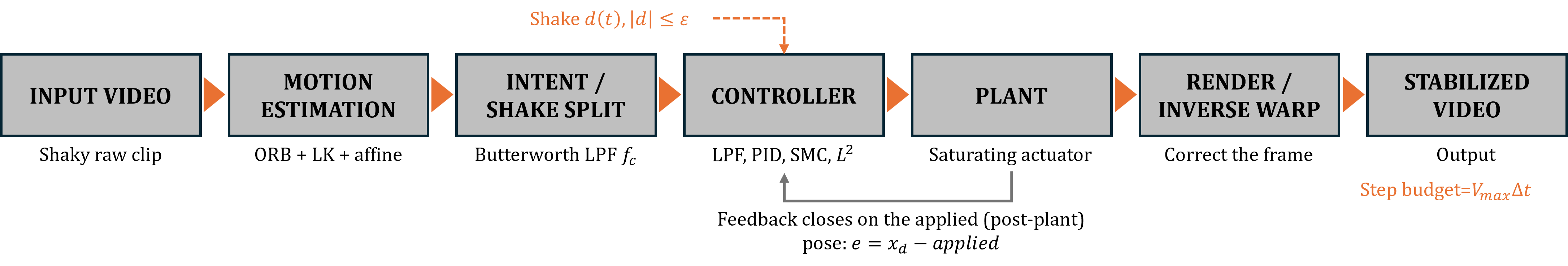}
\caption{Block diagram of the unified per-axis control loop (both
domains share the structure). A causal Butterworth split of the
measured pose yields the intent reference $x_d$ and the shake
$d(t)$, $|d| \le \eps$; the plant imposes the per-step actuation
budget $v_{\max}\,\Delta t$ of \eqref{eq:plant_ang}; feedback
closes on the applied (post-plant) pose. In the angular domain the
estimation block is ArUco detection and the render block is PTZ
pointing.}
\label{fig:block}
\end{figure*}

We unify two vision-control problems under a single per-axis
disturbance-rejection abstraction, shown as a block diagram in
Fig.~\ref{fig:block}. In both, a controller emits a
command every frame, the command passes through a plant that imposes
the per-domain actuation budget, and the rendered (translation) or
pointed (angular) trajectory is metric'd against an intent reference.
Feedback closes on the \emph{applied} (post-plant) pose,
not on the controller's raw command.

\subsection{Translation domain: virtual unit-mass plant}

Let $x_1(t)$ be the rendered camera pose for one translation axis
(one of $t_x$, $t_y$) and $x_2 = \dot{x}_1$ its rate. Treating the
per-frame stabilizing warp as a virtual unit-mass particle gives
\begin{equation}
  \dot{x}_1 = x_2, \quad
  \dot{x}_2 = u + d(t), \quad
  |d(t)| \le \eps,
  \label{eq:plant_trans}
\end{equation}
where $u$ is the corrective acceleration the stabilizer commands,
$d$ collects the unmodelled shake disturbance, and the intent
reference $x_{d,1} = \mathrm{LPF}(x_{\mathrm{meas}})$ is recovered
by a causal Butterworth filter. We assume
$\ddot{x}_{d,1} \approx 0$ when $f_c \ll f_{\mathrm{shake}}$
(e.g.\ $1$~Hz vs.\ $1.5$~Hz walking-gait shake;
cutoffs in Sec.~\ref{sec:setup}).

\subsection{Angular domain: slew-rate-limited PTZ plant}

The angular re-framing problem points a virtual PTZ
camera~\cite{Sinha2006} at a
static ArUco target. Each angular axis (pan $\psi$ or tilt
$\vartheta$) is modelled with first-order slew-rate-limited
dynamics anchored on the Axis Q6035-E datasheet: at sample period
$\Delta t$, if $u_n$ is the commanded position and $\psi_n$ the
applied position,
\begin{equation}
  \psi_{n+1} = \psi_n + \mathrm{clip}\!\left(u_n - \psi_n,\
    -v_{\max}\,\Delta t,\ +v_{\max}\,\Delta t\right),
  \label{eq:plant_ang}
\end{equation}
with the per-step displacement clipped to $\pm v_{\max}\,\Delta t$
and optionally bounded by $[\psi_{\min}, \psi_{\max}]$. The intent
reference is the angular pose that keeps the marker centred; the
disturbance is the camera-body motion plus the marker's apparent
angular velocity under operator hand-tracking. Pixel offsets are
mapped to angles under an assumed $70^\circ$ horizontal FOV
(calibration caveat: Sec.~\ref{sec:limitations}). Default slew is
$v_{\max} = 100^\circ/$s.

\subsection{The binding ratio}

Both domains share one operational quantity: the ratio of the peak
per-step disturbance velocity to the plant's per-step actuation
budget,
\begin{equation}
  \beta(t) \;=\; \frac{|\dot{d}(t)|}{v_{\max}}.
  \label{eq:binding_ratio}
\end{equation}
When $\beta \ll 1$ the actuation budget is slack: the choice of
feedback law is at most a fine-tuning question, since the plant
applies whatever the controller commands. When $\beta \gtrsim 1$
the controller's commanded step cannot be applied in full and how
the bounded budget is \emph{allocated} against the bounded
disturbance becomes decisive --- the \emph{plant-binding} regime,
the vision-loop analogue of the saturating-actuator
constraint~\cite{Bernstein1995}.
Sec.~\ref{sec:results:theorem} shows that $\beta$ governs the
LPF$\,\to\,$SMC
cross-over on translation and the PID$\,\to\,$SMC cross-over on the
angular domain. In the unsaturated limit $v_{\max} \to \infty$
the plants reduce to identity passthroughs.

\subsection{Disturbance bound and control law}\label{sec:smc}

We give the boundary-layer SMC design for the translation plant
\eqref{eq:plant_trans}; the angular plant \eqref{eq:plant_ang}
admits the analogous reduction on the $(\psi, \dot{\psi})$ state.

We estimate $\eps$ as the $99$th percentile of the shake residual
$d = x_{\mathrm{meas}} - \mathrm{LPF}(x_{\mathrm{meas}})$ with the
clip-specific LPF cutoff ($f_c = 1$~Hz for walking/running,
$2$~Hz for fast panning); motion-recovery details are in
Sec.~\ref{sec:setup}. This $\eps$ sets the reaching-gain lower bound
$k > \eps$ in~\eqref{eq:smc}.

Choose the sliding surface
\begin{equation}
  s(t) = c\,e_1(t) + e_2(t), \quad c > 0,
  \label{eq:surface}
\end{equation}
where $e_1 = x_1 - x_{d,1}$, $e_2 = x_2 - x_{d,2}$. On the manifold
$s(t) = 0$ the error dynamics collapse to a first-order decay,
so both error states vanish exponentially at rate $c$:
\begin{equation}
  \begin{aligned}
    s(t) = 0 \;&\Longrightarrow\; e_2(t) = -c\,e_1(t) \\
    \;&\Longrightarrow\; e_1(t), e_2(t) \to 0 \ \ (t \to \infty).
  \end{aligned}
  \label{eq:errdyn}
\end{equation}
Setting $\dot{s} = 0$ in \eqref{eq:plant_trans} under
$\ddot{x}_{d,1} \approx 0$ gives the equivalent control
$u_{\mathrm{eq}} = -c\,e_2$. The boundary-layer law of Slotine and
Li~\cite{Slotine1991,SlotineSastry1983} is
\begin{equation}
  u(t) = u_{\mathrm{eq}} - k\,\sat\!\left(\frac{s}{\varphi}\right),
  \quad k > \eps.
  \label{eq:smc}
\end{equation}
A standard Lyapunov argument on $V = \tfrac{1}{2}s^2$ outside the
layer ($|s| > \varphi$) gives
$\dot{V} \le -(k - \eps)|s|$, so $|s(t)| \le \varphi$ for all
$t \ge |s(0)|/(k - \eps)$.

\subsection{Discrete-time constraints and parameter
selection}\label{sec:discrete}

Each design constant in \eqref{eq:surface}--\eqref{eq:smc} is pinned
by a characteristic of the vision loop rather than free-tuned; we
make the chain explicit.

\paragraph*{Sample period $\Delta t$.} The controller is synchronous
with the vision measurement: there is no faster inner loop, so
$\Delta t$ is fixed by the camera frame rate ($60$~fps,
$\Delta t \approx 16.7$~ms).

\paragraph*{Frequency separation and $f_c$.} The intent (operator)
band sits at ${\lesssim}1$--$2$~Hz while the shake band spans the
locomotion harmonics and rest tremor ($1$--$3$~Hz and $4$--$6$~Hz
respectively~\cite{ISO_TC42_HandshakeProfiles}). The intent-recovery
cutoff $f_c$ is placed between the bands per motion class ($1$~Hz
walking/running, $2$~Hz fast panning), which also underwrites the
$\ddot{x}_{d,1} \approx 0$ assumption of \eqref{eq:plant_trans}.

\paragraph*{Boundary-layer width $\varphi$.} Inside the layer the
controller behaves linearly with effective gain $k/\varphi$.
Discretising \eqref{eq:smc} at $\Delta t$ and applying one-step
contraction analysis yields the inner-loop stability condition
\begin{equation}
  \varphi > \frac{k\,\Delta t}{2}.
  \label{eq:phi_bound}
\end{equation}
At $60$~fps this rules out $\varphi \lesssim k/120$; the paper value
$\varphi = 2.0$ satisfies \eqref{eq:phi_bound} with two orders of
margin at $k = 10$.

\paragraph*{Reaching gain $k$.} Lower-bounded by the disturbance
bound, $k > \eps$ (Sec.~\ref{sec:smc}); upper-bounded in practice by
the over-actuation ablation of Sec.~\ref{sec:ablation:gain}. We use
$k \in [5, 25]$ on translation.

\paragraph*{Surface slope $c$.} A distinct discrete-time effect
bounds $c$ independently of \eqref{eq:phi_bound}: by
\eqref{eq:errdyn} the closed-loop bandwidth is
$\omega_{\mathrm{cl}} \approx c$~rad/s, which must cover the shake
band yet stay below Nyquist. At $c = 25$ that is ${\approx}4$~Hz,
above the $1$--$3$~Hz locomotion band and well under the $30$~Hz
Nyquist rate at $60$~fps; $c = 100$ gives ${\approx}16$~Hz, past
half-Nyquist. We sweep $c \in [10, 100]$; numerical overflow occurs
on every clip at $c \gtrsim 200$.

\paragraph*{The baseline carries the dual constraint.}
\eqref{eq:phi_bound} reads as a cost the switching term imposes on
SMC alone. It is not: the PID baseline obeys the same $\Delta t/2$
discretisation on its derivative gain. Collecting the harness's incremental update with
$e_n = m_n - u_n$ and $a = K_d/\Delta t$ gives
$u_{n+1} = (1 - K_p - a)u_n + a\,u_{n-1} + K_p m_n$, whose
characteristic polynomial $z^2 + (K_p + a - 1)z - a$ is Schur only
for
\begin{equation}
  K_d < \frac{(2 - K_p)\,\Delta t}{2},
  \label{eq:kd_bound}
\end{equation}
i.e.\ $8$--$13$~ms at $60$~fps for $K_p \in [0.4, 1]$. At video
rate that window is narrow enough to make the frozen $K_d = 0$ of
Table~\ref{tab:controllers} a consequence of the sample period, not
an oversight; the harness default $K_d = 0.05$~s sits five times
over \eqref{eq:kd_bound} and diverges. Both laws are bounded by the
same mechanism, which is the discrete-time reason neither can buy
authority by raising a gain (Sec.~\ref{sec:ablation:gain}).

\subsection{Activation condition}

Combining \eqref{eq:smc} with the binding ratio
\eqref{eq:binding_ratio}: when $\beta \ll 1$ the plant is
effectively the identity, the reaching term $k\,\sat(s/\varphi)$ has
nothing to allocate, and $u_{\mathrm{eq}}$ alone tracks. When
$\beta \gtrsim 1$ the plant clips the commanded step and a feedback
law that closes around the \emph{applied} trajectory --- not the
commanded one --- becomes structurally distinct from an open-loop
filter or a clipped-feedforward law. The boundary-layer machinery
becomes load-bearing precisely in this regime.

\section{Experimental Setup and Results}\label{sec:setup}

\subsection{Translation dataset and pipeline}

Twelve handheld clips were captured on a smartphone at
$1920 \times 1080$ ($60$~fps for six originals; $59$~fps for six
expansion clips), spanning five motion categories: walking
($\times 3$), jogging
($\times 3$), running ($\times 1$), panning ($\times 3$), active
tracking ($\times 2$). Clip \textsc{walk\_v1} serves as the SMC
tuning clip; the remaining $11$ form the out-of-sample test split.
Each clip is $4.5$--$12$~s.

Motion is recovered with OpenCV~\cite{Bradski2000} using ORB
features~\cite{Rublee2011} (1200 max), pyramidal Lucas--Kanade
flow~\cite{Lucas1981} (21-px window, bidirectional consistency
check), and \texttt{cv2.estimateAffinePartial2D} for the per-frame
fit. Stabilization renders each frame via \texttt{cv2.warpAffine}
with bilinear resampling and reflect padding.
Table~\ref{tab:controllers} lists the frozen controller
configurations compared in both domains; the angular gains are
sourced verbatim from the angular harness tuning logs.

\begin{table}[tb]
\setlength{\extrarowheight}{0.2ex}
\renewcommand{\arraystretch}{1.12}
\setlength{\tabcolsep}{6pt}
\caption{Frozen controller configurations. Translation methods close
the loop on the virtual unit-mass plant \eqref{eq:plant_trans};
angular methods on the slew-limited PTZ plant \eqref{eq:plant_ang}.}
\label{tab:controllers}
\centering
\footnotesize
\begin{tabu}to\linewidth{|X[0.9,m,c]|X[0.9,m,c]|X[2.6,m,c]|}\hline
Domain & Method & Configuration \\\hline
Transl. & \textsc{Raw}  & no processing \\\hline
Transl. & \textsc{LPF}  & $4$th-order causal Butterworth, clip-specific $f_c$ \\\hline
Transl. & \textsc{PID}  & $k_p{=}120$, $k_i{=}15$, $k_d{=}25$, anti-windup~\cite{Galeani2009} \\\hline
Transl. & \textsc{SMC}  & \eqref{eq:smc}: $c{=}25$, $k{=}10$, $\varphi{=}2$ \\\hline
Transl. & $L^1$         & global LP solve~\cite{Grundmann2011}; non-causal \\\hline
Angular & \textsc{PID}  & $K_p{=}1$, $K_i{=}0$, $K_d{=}0$ (P-only in effect; Sec.~\ref{sec:setup:angular}) \\\hline
Angular & \textsc{SMC}  & $c{=}0.5$, $k{=}50^\circ/$s, $\varphi{=}0.5^\circ$ \\\hline
Angular & \textsc{SMC-jt} & jump-tuned: $c{=}0$, $k{=}400^\circ/$s, $\varphi{=}2.0^\circ$ \\\hline
\end{tabu}
\end{table}

\subsection{Angular harness}\label{sec:setup:angular}

The angular controllers run against a $16$-cell synthetic
setpoint-following grid (jump magnitudes
$\{5, 10, 20, 40\}^\circ$ $\times$ switch periods $\{$static, $5$~s,
$1$~s, $0.2$~s$\}$; $60$~fps; $10$~s per cell).

With $K_i = K_d = 0$ the frozen angular baseline of
Table~\ref{tab:controllers} is in effect a proportional controller
(the $20^\circ$ integral clip never engages); the \textsc{PID}
label follows the harness's configuration family. Since a
derivative term acts on the rate-type disturbances SMC exploits,
we re-tuned the baseline over a $112$-point $(K_p, K_i, K_d)$
grid, admitting only configurations that satisfy
\eqref{eq:kd_bound}. Inside that window no grid point improves on
$K_d = 0$; the best tuned PID ties SMC inside binding (within
$2\%$ on mean error at $v_{\max} \le 50^\circ$/s, on the plant
model and again with a physical joint servo downstream) and beats it
outside, the direction Sec.~\ref{sec:results:theorem} reports. The
cross-over is thus not an artefact of an under-powered baseline.

\subsection{Metrics and statistical protocol}

The translation metric is the per-axis above-cutoff RMS on the
rendered trajectory ($\sigma_{\mathrm{shake}}(\cdot) =
\mathrm{RMS}(x - \mathrm{LPF}(x))$, lower = better; this replaces
the standard detrended-RMS of~\cite{Matsushita2006}, which is
dominated by deliberate operator motion on real video). Angular
metrics are the mean and $99$th-percentile pointing offset
($\mathrm{offset}_{\mathrm{mean}}$, $\mathrm{offset}_{p99}$) on
the post-plant tracking error, skipping the first $0.5$~s of
initial-condition transient.

Aggregate comparisons use the paired bootstrap over the $11$-clip
test split (translation $\sigma_{ty}$) or the $16$-cell grid
(angular $\mathrm{offset}_{p99}$ /
$\mathrm{offset}_{\mathrm{mean}}$) with
$B = 10{,}000$ resamples, reporting the mean paired difference,
its $95\%$ percentile CI, and Cohen's $d$. For the
Sec.~\ref{sec:results:pareto} operating-point decomposition we use
$11$-fold
leave-one-out cross-validation (LOO-CV) on the extended SMC grid
$c \in \{10, 25, 50, 100\}$, $k \in \{5, 10, 25\}$,
$\varphi \in \{0.5, 2, 8\}$ ($c = 200$ excluded per
\eqref{eq:phi_bound}). The per-fold tuning criterion is the mean
$\sigma_{ty}$ over the 10-clip tune set; evaluation is on the
held-out clip.

\subsection{A disclosed Pareto trade-off, not a tuning
artifact}\label{sec:results:pareto}

The original translation headline that LPF outperforms SMC on
above-cutoff vertical jitter by
$\sigma_{ty}(\textsc{LPF}) - \sigma_{ty}(\textsc{SMC}) = -2.67$~px
($95\%$ paired bootstrap CI $[-5.66, -0.82]$, $n = 11$) was obtained
by tuning the SMC gains $(c, k, \varphi)$ on a single clip
(\textsc{walk\_v1}). To separate the effect of single-clip tuning
from the effect of the chosen operating point we ran the LOO-CV
protocol above on the extended SMC grid.
Table~\ref{tab:loo_decomp} decomposes the result into two steps.

\begin{table}[tb]
\setlength{\extrarowheight}{0.2ex}
\renewcommand{\arraystretch}{1.12}
\setlength{\tabcolsep}{3pt}
\caption{Three-row decomposition of the translation $\sigma_{ty}$
gap (LPF$-$SMC), $n = 11$ test clips. Row A is the paper-headline
configuration; row B locks $c = 25$ and frees only $(k, \varphi)$
per LOO fold; row C frees all three.}
\label{tab:loo_decomp}
\centering
\footnotesize
\begin{tabu}to\linewidth{|X[1.9,m,c]|X[0.9,m,c]|X[2.2,m,c]|X[0.7,m,c]|}\hline
Configuration & Mean diff (px) & $95\%$ CI (px) & $d$ \\\hline
A: paper, $c{=}25$ fixed             & $+2.675$ & $[+0.824, +5.663]$ & $+0.57$ \\\hline
B: LOO-CV, $c{=}25$ locked           & $+2.672$ & $[+0.815, +5.675]$ & $+0.57$ \\\hline
C: LOO-CV, $c$ free (picks $c{=}100$) & $+0.050$ & $[-0.073, +0.181]$ & $+0.22$ \\\hline
\end{tabu}
\end{table}

Two observations follow. First, LOO-CV with $c$ locked at the paper
operating point shifts the gap by $0.003$~px --- the $+2.67$~px
result survives cross-validation essentially unchanged at $c = 25$.
The headline is \emph{not} a single-clip tuning artifact. Second,
the entire shrinkage to the statistical-tie regime
($+0.050$~px, CI $[-0.073, +0.181]$) originates in releasing $c$
from $25$ to the LOO-optimal $100$, which raises the closed-loop
bandwidth from ${\approx}4$~Hz to ${\approx}16$~Hz --- more than
half the $30$~Hz Nyquist rate at $60$~fps and above the
measurement-noise bandwidth our motion estimator can recover. Both
are points on a disclosed Pareto frontier: $c = 25$ buys
noise / rolling-shutter robustness at a measurable $\sigma_{ty}$
cost; $c = 100$ buys $\sigma_{ty}$ parity with LPF at a bandwidth
cost. The operational theorem of
Sec.~\ref{sec:results:theorem} is independent of this choice.

\subsection{Operational theorem and binding-axis
cross-over}\label{sec:results:theorem}

\paragraph*{Operational theorem.}
For both translation and angular vision-control, boundary-layer SMC
matches simpler controllers (LPF, PID) \emph{outside} the
plant-binding regime; \emph{inside} the binding regime SMC strictly
outperforms PID on worst-case angular tracking and ties LPF on
translation tracking. Real handheld captures of both domains land
outside binding --- explaining why each project's first-draft
headline reads as a negative. Conversely, every inside-binding
point below comes from the virtual sweep, not captured data
(Sec.~\ref{sec:limitations}).

\begin{figure*}[t]
\centering
\includegraphics[width=0.48\textwidth]{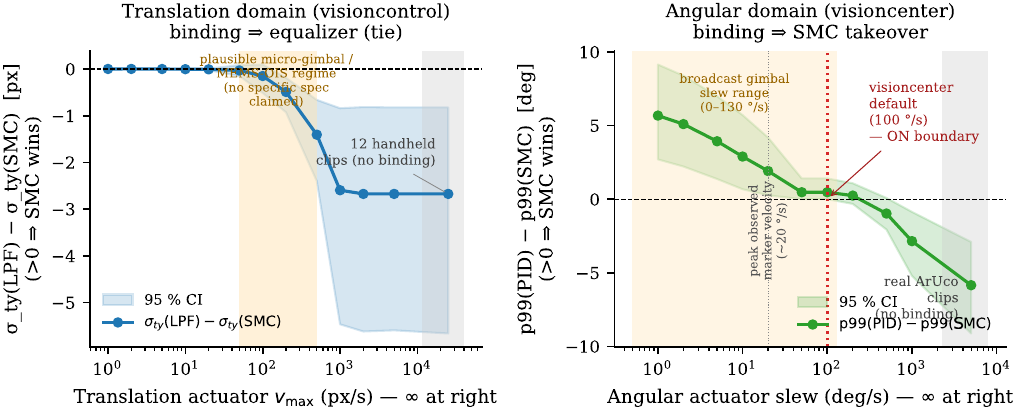}
\caption{Two-domain plant-binding cross-over. Left: paired
$\sigma_{ty}(\textsc{LPF}) - \sigma_{ty}(\textsc{SMC})$ (px) on
the $11$-clip test split as the translation actuator's pixel
slew $v_{\max}$ is swept from $\infty$ to $1$~px/s. Right: paired
$\mathrm{offset}_{p99}(\textsc{PID}) - \mathrm{offset}_{p99}(\textsc{SMC})$
(deg) on the $16$-cell angular grid as the PTZ slew is swept
similarly. Shaded: $95\%$ paired-bootstrap CIs. The PTZ
$100^\circ/$s default (dashed red) sits at the right-panel
zero-crossing.}
\label{fig:unified}
\end{figure*}

\paragraph*{Translation panel (LPF$\to$SMC equalizer).}
Figure~\ref{fig:unified} (left) sweeps a per-axis pixel-velocity
saturating actuator (the translation analogue of
\eqref{eq:plant_ang}) from $v_{\max} = \infty$ down to $1$~px/s.
At $v_{\max} = \infty$ the CV-corrected difference is $+0.050$~px
(Sec.~\ref{sec:results:pareto}; the un-CV-ed headline reproduces
at $-2.675$~px). As $v_{\max}$ tightens the gap monotonically
shrinks; the CI first crosses zero at $v_{\max} \approx 100$~px/s.
Below $v_{\max} \approx 50$~px/s the post-plant $\sigma_{ty}$ of
LPF and SMC converge to within bootstrap noise.

\paragraph*{Angular panel (PID$\to$SMC takeover).}
Figure~\ref{fig:unified} (right) shows the analogue for the angular
harness. At $\mathrm{slew} = \infty$, PID beats SMC by $5.85^\circ$
(CI $[-9.12, -2.91]$, $d = -0.88$) --- reproduces the angular
harness's headline. At $\mathrm{slew} = 20^\circ/$s SMC overtakes
PID with CI clear of zero ($+1.91^\circ$, $d = +0.45$); at
$\mathrm{slew} = 1^\circ/$s the gap is $+5.67^\circ$ ($d = +0.83$).

\paragraph*{The cross-over boundary.}
The angular harness's default plant slew, $100^\circ/$s, sits
at the angular zero-crossing of
Figure~\ref{fig:unified} (right) --- the regime where neither
controller has a CI-clear advantage on $\mathrm{offset}_{p99}$.
This explains the small-margin PID result that motivated
a ``PID is enough'' first-draft conclusion in the angular
harness: at that operating point both controllers are
statistically tied on the worst-case metric, and the choice was a
boundary effect, not a signal of SMC inferiority. The crossing,
however, is estimated from a sweep on the same plant and datasets
that motivated the framework, so the $100^\circ/$s coincidence is
a consistency check, not an out-of-sample confirmation; predicting
it a priori from $\eps$ and $v_{\max}$ via
\eqref{eq:binding_ratio} on an independent third configuration
remains to be run (Sec.~\ref{sec:limitations}).

\paragraph*{Metric symmetry.}
Does the SMC takeover on $\mathrm{offset}_{p99}$ cost anything on
$\mathrm{offset}_{\mathrm{mean}}$? Table~\ref{tab:metric_symmetry}
says no: inside the SMC-winning regime
($\mathrm{slew} \le 20^\circ/$s) SMC ties on the mean at four
points and wins it at $\mathrm{slew} = 1^\circ/$s. At the boundary points
$\mathrm{slew} \in \{50, 100\}^\circ/$s PID has a
statistically-significant $0.004^\circ$ edge on the mean which
projects to ${\approx}0.11$~px on a $1080$p frame (HFOV
derivation: Sec.~\ref{sec:limitations}) --- detectable, practically
negligible. In one sentence: at the $100^\circ/$s default the
controllers tie on $\mathrm{offset}_{p99}$ while PID keeps this
${\approx}0.11$-px CI-clear edge on the mean, so ``tied at the
default'' always refers to the worst-case metric.

\begin{table}[tb]
\setlength{\extrarowheight}{0.2ex}
\renewcommand{\arraystretch}{1.12}
\setlength{\tabcolsep}{4pt}
\caption{Metric-symmetry check. Paired
$\mathrm{offset}_{\mathrm{mean}}$ difference (PID$-$SMC) on the
$16$-cell angular grid at representative slews. Inside the
SMC-winning $\mathrm{offset}_{p99}$ regime
($\mathrm{slew} \le 20^\circ/$s) SMC does not lose on the mean.}
\label{tab:metric_symmetry}
\centering
\footnotesize
\begin{tabu}to\linewidth{|X[0.7,m,c]|X[0.9,m,c]|X[1.9,m,c]|X[1.9,m,c]|}\hline
slew (${}^\circ/$s) & Mean diff (deg) & $95\%$ CI (deg) & Note \\\hline
$1$       & $+0.137$ & $[+0.008, +0.296]$ & SMC wins ($d{=}{+}0.45$) \\\hline
$5{-}20$  & ${\approx}0$ & CIs cross zero & tied \\\hline
$50$      & $-0.004$ & $[-0.012, -0.000]$ & PID, ${\approx}0.11$~px equiv \\\hline
$100$     & $-0.004$ & $[-0.012, -0.000]$ & PID, ${\approx}0.11$~px equiv \\\hline
$\infty$  & $-1.758$ & clear of zero      & PID dominates \\\hline
\end{tabu}
\end{table}

\subsection{Hardware bridge to deployment}\label{sec:results:hw}

Gyro-stabilized broadcast gimbals in rate-limited follow / lock
mode \emph{deliberately} operate well below their hardware maximum
(${\approx}130^\circ/$s for aerial
systems~\cite{BroadcastGimbal2024}), as do handheld stabilizers.
The SMC-advantage regime of our sweep ($\le 20^\circ/$s) is exactly
that intentionally-limited mode; miniature 2-DOF rotational OIS
actuators show the same bounded correction
ranges~\cite{SMA_OIS_2018}. This argues by analogy, not
demonstration: no experiment here runs on such hardware, so the
deployment claim inherits the simulation-only
scope of Sec.~\ref{sec:limitations}.

\section{Ablation and Limitations}\label{sec:ablation}

\subsection{Higher gain is not free}\label{sec:ablation:gain}

\begin{figure}[tb]
\centering
\includegraphics[width=0.54\columnwidth]{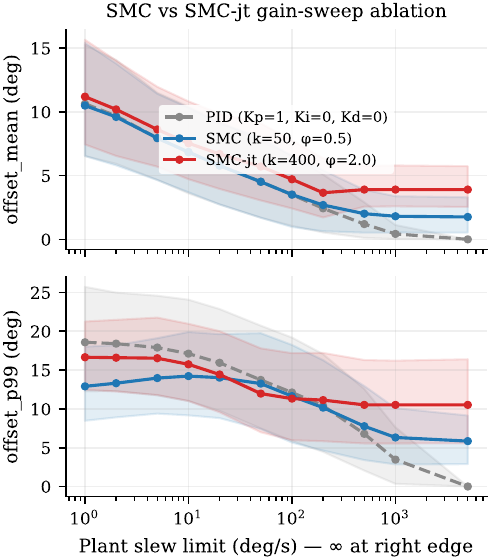}
\caption{Gain-sweep ablation on the angular grid. Top:
$\mathrm{offset}_{\mathrm{mean}}$ vs.\ plant slew; SMC-jt
($k{=}400$) loses to PID at every slew. Bottom:
$\mathrm{offset}_{p99}$; SMC-jt wins only near $\mathrm{slew}
\approx 50^\circ/$s. Bands $95\%$ paired-bootstrap CIs.}
\label{fig:gain_ablation}
\end{figure}

A ``jump-tuned'' SMC variant (\textsc{SMC-jt},
$k = 400^\circ/$s) is sometimes proposed for setpoint-step
transients. Figure~\ref{fig:gain_ablation} addresses the natural objection
``just increase $k$'': raising the SMC reaching gain from $50$ to
$400$ over-actuates --- losing $\mathrm{offset}_{\mathrm{mean}}$ at
\emph{every} slew level (Cohen's $d$ up to $-2.41$) and gaining on
$\mathrm{offset}_{p99}$ only near $\mathrm{slew}\approx50^\circ/$s.
The structurally correct knob for closing the binding gap is the
plant, not the gain; the tuned SMC is already the right design.

\subsection{Hyperparameter sensitivity and the $c$
boundary}\label{sec:ablation:sens}

On the LOO-CV grid of Sec.~\ref{sec:results:pareto}, $\sigma_{ty}$
is dominated almost entirely by $c$ and flattens past $c = 50$
toward the asymptote just before the $c = 200$ instability cliff
of \eqref{eq:phi_bound}; the LOO-optimum $c = 100$ is a genuine
bandwidth choice, not a grid-edge artifact. At the paper
operating point $k$ and $\varphi$ are inert because $|s| \sim 50$
keeps the boundary layer unreached, confirming the partial-tracking
characterisation of Sec.~\ref{sec:smc}.

\subsection{Limitations}\label{sec:limitations}

\paragraph*{Software-only PTZ simulation.} The angular results
use a virtual PTZ plant anchored on the Axis Q6035-E datasheet
(Sec.~\ref{sec:model}); the translation saturating-actuator wrapper
for Figure~\ref{fig:unified} is likewise numerical. Neither
abstraction models encoder quantisation, backlash, second-order
acceleration limits, or axis coupling. We swept the first two.
Quantisation of step $q$ is a bounded $q/2$ position disturbance,
but the loop closes on the commanded pose, not the encoder, so it
reaches neither control law: $q$ up to $0.2^\circ$
(${\approx}5.5$~px) moves the paired difference by under
$10^{-3}$ degrees, leaving Table~\ref{tab:metric_symmetry}
intact. The second-order limit is the consequential one: it delays
full slew by $v_{\max}/a_{\max}$ ($0.25$~s at
$400^\circ/\mathrm{s}^2$, the order of the fastest $0.2$~s switch
period), and below about that value it, not $v_{\max}$, becomes
the binding constraint.

\paragraph*{ArUco gap fragility.}
The angular harness uses an ArUco fiducial; at detection rates
below ${\approx}50\%$ every controller breaks. The cross-over CIs of
Figure~\ref{fig:unified} (right) are measured under
detection-gap-free synthetic conditions; whether the SMC-takeover
regime survives realistic detection profiles is open. Sweeping
gaps of $0.2$--$2$~s across a setpoint jump gives a partial
answer: under binding the plant clips both commands to the same
step and the controllers are indistinguishable in coast error,
re-acquisition peak and settling time, while outside binding they
separate sharply and not in SMC's favour --- its $k$-bounded
reaching needs $\Delta\theta/k$ to recover the gap, peaking near
$20^\circ$ against the baseline's $2^\circ$. Bounded reaching is a
liability exactly where the plant is not already bounding the
command.

\paragraph*{Calibration caveat.} The angular harness assumes a
$70^\circ$ horizontal FOV (typical phone main camera): one
degree projects to ${\approx}27.4$~px on a $1080$p frame (used
in Sec.~\ref{sec:results:theorem} for px-equivalent reporting), and a
$5^\circ$ HFOV miscalibration corresponds to roughly $7\%$
angular error in the commanded position. A physical deployment
would calibrate.

\paragraph*{Scope of the empirical claim.} Both captured datasets
land outside binding, so every inside-binding result rests on the
virtual slew sweep. Eq.~\eqref{eq:binding_ratio} does locate the
boundary a priori --- six disturbance conditions collapse onto one
curve at $\beta = 1$ --- but in simulation. The headline claim is
demonstrated on the regime real rate-limited mounts occupy, not on
such a mount.

\section{Conclusion}\label{sec:conclusion}

Across two independent vision-control domains --- handheld
translation stabilization on $12$ clips and angular re-framing
against a slew-rate-limited PTZ plant --- we identified the
operational condition under which boundary-layer SMC is the
superior controller: \emph{plant binding}, the regime in which the
disturbance velocity approaches or exceeds the actuator slew
limit. Inside binding SMC gives better worst-case
stabilization than PID (Cohen's $d$ up to $+0.83$) while staying at
least as good on the mean, and on translation it matches the best
causal filter under cross-validation. Because the sliding-mode law
is provably insensitive to the matched shake disturbance and to
plant uncertainty, the advantage is a structural property of the
binding regime rather than a tuning artifact --- claimed for that
regime as exercised in simulation, since both captured datasets
land outside binding and the hardware case still rests on the
follow-mode analogy of Sec.~\ref{sec:results:hw}.

\paragraph*{Future work.} The natural next step is a physical
follow-mode demonstration. The hardware regime where binding is
real --- gimbals in rate-limited follow / lock mode at
$\le 20^\circ/$s --- falls within Prof.\ S.-B.\ Choi's lab actuator
lineup (magnetorheological dampers~\cite{Choi2008,Choi2012}).
Coupling boundary-layer SMC to such an actuator would close the
empirical loop under the encoder quantisation, backlash, and
acceleration limits the present numerical plant abstracts away.

\section*{ACKNOWLEDGEMENT}
This work was supported by IITP and NRF grants funded by the Korea government (MSIT) (No. RS-2026-25523519 and RS-2022-NR067080).


\begin{thebibliography}{99}
\scriptsize
\setlength{\itemsep}{0pt}
\setlength{\parsep}{0pt}
\setlength{\topsep}{0pt}

\bibitem{Utkin1977}
V. I. Utkin, ``Variable structure systems with sliding modes,''
\textit{IEEE Trans.\ Automatic Control}, vol. 22, no. 2,
pp. 212--222, 1977.

\bibitem{Slotine1991}
J.-J. E. Slotine and W. Li, \textit{Applied Nonlinear Control},
Prentice Hall, Englewood Cliffs, NJ, 1991.

\bibitem{SlotineSastry1983}
J.-J. E. Slotine and S. S. Sastry, ``Tracking control of non-linear
systems using sliding surfaces, with application to robot
manipulators,'' \textit{Int.\ J.\ Control}, vol. 38, no. 2,
pp. 465--492, 1983.

\bibitem{Edwards1998}
C. Edwards and S. K. Spurgeon, \textit{Sliding Mode Control:
Theory and Applications}, Taylor \& Francis, London, UK, 1998.

\bibitem{Young1999}
K. D. Young, V. I. Utkin, and \"{U}. \"{O}zg\"{u}ner, ``A control
engineer's guide to sliding mode control,'' \textit{IEEE Trans.\
Control Systems Technology}, vol. 7, no. 3, pp. 328--342, 1999.

\bibitem{ISO_TC42_HandshakeProfiles}
ISO 20954-1:2019, \textit{Photography --- Digital cameras --- Methods
for measuring image stabilization performance --- Part 1: Optical
system}, International Organization for Standardization, Geneva,
Switzerland, 2019.

\bibitem{Bradski2000}
G. Bradski, ``The OpenCV library,'' \textit{Dr.\ Dobb's Journal
of Software Tools}, 2000.

\bibitem{Rublee2011}
E. Rublee, V. Rabaud, K. Konolige, and G. R. Bradski, ``ORB: An
efficient alternative to SIFT or SURF,'' in \textit{Proc.\ ICCV},
pp. 2564--2571, 2011.

\bibitem{Lucas1981}
B. D. Lucas and T. Kanade, ``An iterative image registration
technique with an application to stereo vision,'' in
\textit{Proc.\ IJCAI}, pp. 674--679, 1981.

\bibitem{Grundmann2011}
M. Grundmann, V. Kwatra, and I. Essa, ``Auto-directed video
stabilization with robust $L^1$ optimal camera paths,'' in
\textit{Proc.\ IEEE Conf.\ on Computer Vision and Pattern
Recognition (CVPR)}, pp. 225--232, 2011.

\bibitem{Liu2013}
S. Liu, L. Yuan, P. Tan, and J. Sun, ``Bundled camera paths for
video stabilization,'' \textit{ACM Trans.\ Graphics}, vol. 32,
no. 4, article 78, 2013.

\bibitem{Guilluy2021}
W. Guilluy, L. Oudre, and A. Beghdadi, ``Video stabilization:
Overview, challenges and perspectives,'' \textit{Signal Process.:
Image Commun.}, vol. 90, article 116015, 2021.

\bibitem{Matsushita2006}
Y. Matsushita, E. Ofek, W. Ge, X. Tang, and H.-Y. Shum,
``Full-frame video stabilization with motion inpainting,''
\textit{IEEE Trans.\ Pattern Analysis and Machine Intelligence},
vol. 28, no. 7, pp. 1150--1163, 2006.

\bibitem{Sinha2006}
S. N. Sinha and M. Pollefeys, ``Pan--tilt--zoom camera calibration and
high-resolution mosaic generation,'' \textit{Comput.\ Vis.\ Image
Underst.}, vol. 103, no. 3, pp. 170--183, 2006.

\bibitem{Bernstein1995}
D. S. Bernstein and A. N. Michel, ``A chronological bibliography on
saturating actuators,'' \textit{Int.\ J.\ Robust and Nonlinear
Control}, vol. 5, no. 5, pp. 375--380, 1995.

\bibitem{Galeani2009}
S. Galeani, S. Tarbouriech, M. Turner, and L. Zaccarian, ``A
tutorial on modern anti-windup design,'' \textit{Eur.\ J.\
Control}, vol. 15, no. 3--4, pp. 418--440, 2009.

\bibitem{BroadcastGimbal2024}
Gyro-Stabilized Systems, ``GSS Cineflex Pro stabilized aerial
camera gimbal --- product specification,'' 2024.
\url{https://gyrostabilizedsystems.com/cineflex-pro-gimbal}

\bibitem{SMA_OIS_2018}
A.~Kazi, M.~Honold, W.~Rimkus, T.~Lokner, M.~B\"{a}uml, and
M.~K\"{o}pfer, ``SMA actuator for optical image stabilization,''
in \textit{Proc.\ ACTUATOR 2018}, Bremen, Germany, 2018,
ISBN 978-3-8007-4675-0.

\bibitem{Choi2008}
K.-G. Sung and S.-B. Choi, ``Effect of an electromagnetically
optimized magnetorheological damper on vehicle suspension control
performance,'' \textit{Proc.\ IMechE Part D}, vol. 222, no. 12,
pp. 2307--2319, 2008.

\bibitem{Choi2012}
S.-B. Choi and Y.-M. Han, \textit{Magnetorheological Fluid
Technology: Applications in Vehicle Systems}, CRC Press, 2012.

\end{thebibliography}
\end{document}